# A novel method of fuzzy time series forecasting based on interval index number and membership value using support vector machine

*Kiran Bisht*
Dept. of Mathematics, Statistics and Computer Science
G. B. Pant University of Agriculture and Technology
Pantnagar, Uttarakhand, 263145, India
kiranbisht96@gmail.com

*Arun Kumar*
Dept. of Mathematics, Statistics and Computer Science
G. B. Pant University of Agriculture and Technology
Pantnagar, Uttarakhand, 263145, India
arun_pal1969@yahoo.co.in

**Abstract:** Fuzzy time series forecasting methods are very popular among researchers for predicting future values as they are not based on the strict assumptions of traditional time series forecasting methods. Non-stochastic methods of fuzzy time series forecasting are preferred by the researchers as they provide more significant forecasting results. There are generally, four factors that determine the performance of the forecasting method (1) number of intervals (NOIs) and length of intervals to partition universe of discourse (UOD) (2) fuzzification rules or feature representation of crisp time series (3) method of establishing fuzzy logic rule (FLRs) between input and target values (4) defuzzification rule to get crisp forecasted value. Considering the first two factors to improve the forecasting accuracy, we proposed a novel non-stochastic method fuzzy time series forecasting in which interval index number and membership value are used as input features to predict future value. We suggested a simple rounding-off range and suitable step size method to find the optimal number of intervals (NOIs) and used fuzzy c-means clustering process to divide UOD into intervals of unequal length. We implement support vector machine (SVM) to establish FLRs. To test our proposed method we conduct a simulated study on five widely used real time series and compare the performance with some recently developed models. We also examine the performance of the proposed model by using multi-layer perceptron (MLP) instead of SVM. Two performance measures RSME and SMAPE are used for performance analysis and observed better forecasting accuracy by the proposed model.

1. Introduction

Forecasting future values is very common in different domains of life. Forecasting based on past observed values is known as time series forecasting. Both stochastic and non-stochastic methods are employed by researchers for time series forecasting in past years. Stochastic methods like moving average (MA), autoregressive moving average (ARMA), vector regression and exponential moving average based models have limitations in handling complex and highly uncertain real world forecasting problems. Due to these limitations non-stochastic methods are preferred over stochastic methods. Fuzzy time series forecasting (FTSF) methods are highly popular in the research field because of linguistic representation these methods more closely illustrate real world scenarios and generally give better results than traditional methods.

Fuzzy set theory was firstly proposed by Zadeh (1965). Based on fuzzy sets, Song and Chissom (1993a, b, 1994) were the first to introduce fuzzy time series forecasting models. These models were based on min-max composition operation having cumbersome computational process which was improved by Chen (1996) using simple arithmetic operations. The whole process of fuzzy time series forecasting has four major steps as follows:

1. Determining universe of discourse (UOD), number of intervals (NOIs) and length of intervals to divide UOD.
2. Obtaining fuzzy sets and fuzzifying the crisp time series to fuzzy time series.
3. Establishing fuzzy logic relationship on fuzzified time series.

4. Defuzzifying fuzzified forecasted value to get crisp output.

Huarng(2001a) examined that length of intervals to decompose UOD determines the performance of forecasting. Huarng (2001a, b) proposed average and distribution-based methods to obtain the effective fixed length of interval and improved the performance of the model by heuristic approach. Afterward many researchers discover new methods to find effective length to divide UOD into intervals of equal length. Li and Cheng (2007) conclude that more accuracy in forecasting can be achieved by taking shorter length of intervals. But real time series are mostly non-uniformly distributed that encouraged researchers to develop methods to partition UOD into unequal intervals. Huarng and Yu (2006) suggested a ratio-based method to determine the length of interval. They partition the UOD into unequal length of intervals and verify the improved performance by empirical analysis. Chen and Chung (2006) proposed a genetic algorithm to divide UOD in unequal length of intervals and forecast higher order fuzzy time series. Yu (2005) refine the lengths of interval during the formulation of fuzzy logic relation. Chen and Phuong (2017), Kuo et al. (2009,2010), Hsu et al. (2010), Huang et al. (2011) used particle swarm optimization technique, Cai et al. (2015) suggest ant colony optimization, Enayatifar et al. (2013), Sadaei et al. (2016) proposed imperialist competitive algorithm for determining the lengths of interval. Fuzzy clustering algorithms like fuzzy c-means, Gustafson-Kessel are also used by researchers for the first step of forecasting algorithms. Cheng et al. (2008), Li et al. (2008, 2011), Liu et al. (2010), Pattanayak et al. (2020) are some researchers applied FCM and Egrioglu et al. (2011) used Gustafson-Kessel clustering algorithm and used membership values directly for forecasting.

Now, coming to the second step of forecasting algorithm, mostly researchers used linguistic values to fuzzify the crisp time series according to the index of intervals obtained by UOD partitioning. Some used triangular and trapezoidal membership function (Cheng at al. (2006), Rubio et al. (2016, 2017)). Yolcu et al (2016) used fuzzy membership values obtained by fuzzy c-means clustering and proposed intersection method for establishing fuzzy logic relations.

After fuzzification of crisp time series next step is to establish fuzzy logic relationships (FLRs). Choosing a method to establish FLRs is a crucial step of the FTSF algorithm, it directly affects the forecasting performance. Distinct methods such as fuzzy relation matrix, fuzzy logic relation groups (FLRGs), artificial neural network (ANN) and recently machine learning algorithms have been employed by the researchers for formulating the rules of fuzzy logic relation. Song and Chissom (1993, 1994) applied min-max composition based fuzzy relation matrix, Aladag et al. (2012) used PSO based fuzzy relation matrix and Wong et al. (2010) used markov chain method for fuzzy relationship matrix. These methods have a cumbersome computation process, so the alternative mostly used for establishing FLRs is FLRGs. Chen (1996), Singh and Borah (2013), Ye et al. (2016),Cheng et al. (2016), Huarng (2001a, 2001b) are some of the researchers that had used FLRGs. In recent years, mostly researchers have been using various kinds of artificial neural network techniques as they are computationally convenient and also state-of-the-art methods. Huarng and Yu (2006) was the first to use back propagation neural network to solve the problem of non-linearity and verify the method by forecasting taiex for the years 1991-2003. Then Aladag et al. (2009), Egrioglu et al. (2009), Yolcu et al. (2016) and Singh and borah (2013) applied feed forward neural networks to train the models for forecasting. Bas et al. (2018) applied Pi-Sigma neural network for determining high order fuzzy logic relations and used particle swarm optimization to optimize the parameters of Pi-Sigma neural network. Panigrahi and Bahera (2018) introduced generalized regression neural network to overcome the problems of traditional ANN and test the proposed model on thirteen real time series data. In 2020, Panigrahi and Bahera was the first implemented machine learning techniques for establishing FLRs. They suggested a modified average based method to find the effective length of interval and used three machine learning techniques namely deep belief network (DBN), long-short term memory (LSTM) and support vector machine (SVM) for modeling FLRs. Pattanayak et al. (2020) developed a new method of FTSF by using membership values with the data and used support vector machine for establishing FLRs. Motivated by the method suggested by Pattanayak et al. (2020) in this paper we proposed a new algorithm of FTSF in which uses interval index number and membership value of the data, also discuss a basic method to find optimum number of intervals (NOIs) and use FCM for determining unequal intervals.

The rest of the paper is structured as follows: Some definitions of FTSF terminologies, short notes on FCM and SVM are given in Section (2) as preliminary discussion. The detailed explanation of proposed method is given in Section (3). A simulated study is conducted in Section (4). The detailed analysis of the performance is carried out in Section (5). At last, in Section (6) we conclude the paper with some future indications.

## 2. Preliminaries

### 2.1 Fuzzy Time Series Forecasting (FTSF)

Definitions of some terminologies used in FTSF are given below:

**Definition 1.** A set $U$ containing all elements of time series can be defined as universe of discourse. It could be discrete or continuous. If $U$ is continuous say $U = [a, b]$ then $a$ and $b$ can be taken as the lower bound and upper bound of time series respectively.

**Definition 2.** Let $U$ (universe of discourse) be a discrete and finite set such that $U = \{x_1, x_2, \ldots x_n\}$. A fuzzy set $A$ of $U$ is expressed as:

$$A = f_A(x_1)/x_1 + f_A(x_2)/x_2 + \ldots + f_A(x_n)/x_n$$

where $f_A$ is the membership function of fuzzy set $A$, $f_A: U \to [0,1]$ and $f_A(x_i)$ is the degree of membership of element $x_i$ in the fuzzy set $A$.

For continuous $U$ fuzzy set $A$ of $U$ is expressed as:

$$A = \int f_A(x_i)/x_i \quad (i = 1,2 \ldots n)$$

**Definition 3.** Fuzzy time series $F(t)$ is a collection of fuzzy sets $f_i(t)(i = 1,2,3 \ldots n)$ defined on $Y(t)(t = 0,1,2,3 \ldots)$ where $Y(t)$ is universe of discourse containing real numbers.

**Definition 4.** Suppose $F(t)$ is caused by $F(t-1)$ then the relation $R(t, t-1)$ between $F(t)$ and $F(t-1)$ called fuzzy logic relation (FLR) which is defined as $F(t) = F(t-1)°R(t, t-1)$ where ° is an arithmetic operator. This type of FLR can represented as $F(t-1) \to F(t)$ and FTS with such relation called first order fuzzy time series.

**Definition 5.** If $F(t)$ is determined by $F(t-1), F(t-2), \ldots F(t-n)$, then FLR is defined as $F(t-n) \ldots F(t-2) F(t-1) \to F(t)$ and the FTS is called $n^{th}$ order fuzzy time series.

### 2.2 Fuzzy c-Means Clustering (FCM)

Fuzzy c-means (FCM) is a clustering algorithm developed by Bezdek (1981). Suppose $X = \{x_1, x_2, \ldots x_n\}$ is the set of data points which have to partition into $2 \leq c \leq n$ clusters. In FCM process, dataset is partitioned into c clusters, the membership value $(u_{ij})$ of each data point $(x_i)$ to each cluster $(j = 1,2, \ldots c)$ and center of the clusters $(v_j)$ are calculated by equation (1) and (2) respectively:

$$u_{ij} = \frac{\left(1/d(x_i, v_j)^2\right)^{\frac{1}{p-1}}}{\sum_{k=1}^{c}(1/d(x_i, v_k)^2)^{\frac{1}{p-1}}} \tag{1}$$

$$v_j = \frac{\sum_{i=1}^{n} u_{ij}^p * x_i}{\sum_{i=1}^{n} u_{ij}^p} \qquad (2)$$

where $d(x_i, v_j)$ denotes euclidean distance between data point $x_i$ and cluster center $v_j$ and $p \in [1, \infty[$ parameter of fuzziness.

$u_{ij}$'s and $v_j$'s are calculated iteratively in order to minimize sum squared error (SSE)

$$SSE = \sum_{i=1}^{n} \sum_{j=1}^{c} u_{ij}^p d^2(x_i, v_j) \qquad (3)$$

subject to the following constraints:

$$u_{ij} \in [0,1]; \quad \sum_{j=1}^{c} u_{ij} = 1; \quad \sum_{i=1}^{n} u_{ij} < n$$

Minimization of SSE ensures the each data point belongs to cluster whose center is nearest to it and it also makes well separated clusters.

### 2.3 Support Vector Machine (SVM)

The use support vector machine for regression task was first documented by Vapnik(1995). This regression technique is known as support vector regression (SVR). Let $\chi = \{(X_1, y_1), (X_2, y_2), \ldots (X_n, y_n)\}$; $X_i \in \mathcal{R}^m, y_i \in \mathcal{R}$ be the training set. In SVR we try to find the best fit function say $f(X)$ which predict $y$ for given $X$ with following two assumptions first is that the error between predicted $y_i$'s and actual values must be less than $\epsilon$ and $f(X)$ must be as flat as possible. The function $f(X)$ is defined as follows:

$$f(X) = w . \varphi(X) + b \qquad (4)$$

where $w$ is weight vector, $b$ is bias and $\varphi(X)$ is a non-linear function. For linear regression we take $X$ in place of $\varphi(X)$ in equation (4).

To ensure the flatness of the function (4), we consider an optimization problem given below:

minimize $\frac{1}{2} w^2$ \qquad (5)

subject to the constraints: $\begin{cases} y_i - w.\varphi(X_i) - b \leq \epsilon \\ w.\varphi(X_i) + b - y_i \leq \epsilon \end{cases}$

In some cases it may difficult to find such regression function that completely fits the training set, may be some training data points fall outside the $\epsilon$ margin. In this case we can reconsider optimization problem defined in equation (5) which find the best fit function $f(X)$ allowing some errors outside the $\epsilon$ margin. Two slack variables $\xi_i$ and $\xi_i^*$ are introduced and the optimization problem became:

minimize $\frac{1}{2} w^2 + C \sum_{i=1}^{l} (\xi_i + \xi_i^*)$ \qquad (6)

subject to the constraints: $\begin{cases} y_i - w.\varphi(X_i) - b \leq \epsilon + \xi_i \\ w.\varphi(X_i) + b - y_i \leq \epsilon + \xi_i^* \\ \xi_i, \xi_i^* \geq 0 \end{cases}$

The parameter $C > 0$ maintains the flatness of the function $f(X)$ and the allowed $\epsilon$ error and also prevents over fitting.

Lagrange dual form of the above optimization problem is easier to compute which is defined as follows:

$$\text{maximize } \left(-\frac{1}{2}\sum_{i=1}^{n}\sum_{j=1}^{n}(\alpha_i - \alpha_i^*)(\alpha_j - \alpha_j^*)K(x_i,x_j) - \epsilon\sum_{i=1}^{n}(\alpha_i - \alpha_i^*) + \sum_{i=1}^{n}y_i(\alpha_i - \alpha_i^*)\right) \quad (7)$$

subject to the constraints: $\begin{cases} \sum_{i=1}^{n}(\alpha_i - \alpha_i^*) = 0 \\ 0 \leq \alpha_i, \alpha_i^* \leq C \end{cases}$

where $\alpha_i$ and $\alpha_i^*$ are non-negative Lagrange multipliers and $K(x_i, x_j)$ is kernel function. We can define different types of kernel such as linear, polynomial as per the regression problem.

### 3. Proposed method

In this section, we proposed a new fuzzy time series forecasting method. We used Fuzzy c-means for partitioning the universe of discourse then utilize both interval index and membership value of crisp data for fuzzification and SVM to determine fuzzy logic relations. The step by step procedure of the method is explained in detail with the well known enrollment data of University of Alabama to understand better.

Step 1. Defining universe of discourse (UOD) for time series data:

Define universe of discourse (UOD) as $U = [y_{min} - d, y_{max} + d]$, where $y_{min}$ and $y_{max}$ are the maximum and minimum values of the real time series and $d$ is a positive integer. For the enrollment data given in table (1) we have $y_{min} = 13055$ and $y_{max} = 19337$ and assuming $d = 8$. So, we have $U = [13047, 19345]$.

Step 2. Clustering the time series data by FCM and generating cluster centers:

Before applying FCM we have to specify the optimal number of clusters($c$) to partition the dataset. Number of cluster effect the performance of forecasting, so identifying optimal number of clusters($c$) is an important task. As our proposed method depend on interval index, keeping this in consideration we roughly round off the range of time series and take number of clusters ($c$) according to the suitable large step size in range.

For example, the enrollment data varies from 13000 to 20000. Taking 1000 as step size, time series roughly clustered in 7 groups. Hence we take $c = 7$ and partition the enrollment data into 7 clusters by FCM. Taking large step size means less number of clusters which avoid overlapping in time series data and cluster them into well separated groups.

The centers of the clusters calculated by FCM for enrollment data are {13573.95, 15130.14, 15448.22, 15885.33, 16825.08, 18190.36 and 19125.23}.

Step 3. Dividing UOD in unequal intervals:

For dividing UOD in unequal intervals we take midpoints of cluster centers as boundaries of intervals. The UOD of enrollment data is divided into 7 unequal intervals given as [13047, 14352.045], [14352.045, 15289.18], [15289.18, 15666.775], [15666.775, 16355.205], [16355.205, 17507.72], [17507.72, 18657.795], [18657.795, 19345].

Step 4. Fuzzifying of the crisp data and obtaining their membership value:

Now we fuzzify each data point to the interval index $i$ which it belongs then find the membership value of the data point in that particular interval by min-max normalization operation. Let, $y$ be the data point lie in the interval [a, b] then its membership value in $i^{th}$ interval is expressed as

$$m = \frac{y-a}{b-a} \quad (8)$$

We take the interval number $i$ and membership value $m$ as features of the data point.

For example the first observation of enrollment data is 13055 which belongs to the first interval [13047, 14352.045], the membership value ($m$) of 13055 in this interval is calculated as

$$m = \frac{13055-13047}{14352.045-13047} = 0.00613$$

So, the features of first observation are $\{1, 0.00613\}$.

Similarly, eighth observation 15861 belongs to the fourth interval [15666.775, 16355.205], the membership value

$$m = \frac{15861-1566.775}{16355.205-15666.775} = 0.28212$$

The features are $\{4, 0.28212\}$.

Step 5. Normalizing the time series and setting an input-output pattern for forecasting:

In this step, first we normalize the time series data by min-max normalization and obtain the normalized time series. Suppose $y$ be an element of crisp time series its normalized value $\|y\|$ is given as $\|y\| = \frac{y - y_{min}}{y_{max} - y_{min}}$ where $y_{min}$ and $y_{max}$ are the minimum and maximum values of real-time series.

Now we set an input-output pattern between the features of crisp time series and normalized time series. The features of the previous $(t-1)$ time step are set as input to get the normalized value of $t$ time step as output.

The input and target values for enrollment data of University of Alabama are shown in Table 2.

Step 6. Using support vector regression (SVR) to establish relation between input and target values:

Now, we use support vector regression to find the fuzzy logic relation between the input features and target values. We train the support vector machine with 80% of data and rest 20% is used for testing the performance. SVM is trained over the training dataset to find the best-fitted function $f(X)$ where, $X \in \mathcal{R}^2$ is feature vector to predict the target value $y$.

Step 7. Obtaining the test results and de-normalizing them:

After training, we feed test features to the trained SVM and obtain the forecasted values. As we have trained SVM over normalized target values so test predicted values are also normalized. To obtain the actual values we de-normalized the forecasted values by min-max de-normalized. Let $\|y\|$ be the normalized forecasted value the actual value $y$ is obtained as follows: $y = \|y\|(b-a) + a$

Step 8. Analyzing the performance of the method:

At last, we use root mean square error (RMSE) and symmetric mean absolute percentage error (SMAPE) to analyze the performance of proposed method.

$$RMSE = \sqrt{\frac{1}{n}\sum_{i=1}^{n}(y_i - y_i')^2} \qquad (9)$$

$$SMAPE = \frac{1}{n}\sum_{i=1}^{n}\frac{|y_i - y_i'|}{(|y_i| + |y_i'|)/2} \qquad (10)$$

Table 1. Enrollment data of University of Alabama

| Year | Actual enrollment | Year | Actual enrollment |
|---|---|---|---|
| 1971 | 13055 | 1982 | 15433 |
| 1972 | 13563 | 1983 | 15497 |
| 1973 | 13867 | 1984 | 15145 |
| 1974 | 14696 | 1985 | 15163 |
| 1975 | 15460 | 1986 | 15984 |
| 1976 | 15311 | 1987 | 16859 |
| 1977 | 15603 | 1988 | 18150 |
| 1978 | 15861 | 1989 | 18970 |
| 1979 | 16807 | 1990 | 19328 |
| 1980 | 16919 | 1991 | 19337 |

Table 2. Input features and target values for enrollment data

| Year | Actual enrollment | Input features | | | | Target values | |
|---|---|---|---|---|---|---|---|
| | | Normalized time series | Forecast year | Interval index | Membership value in the interval | | |
| 1971 | 13055 | 0.0 | 1972 | 1 | 0.00613 | 0.08086 | |
| 1972 | 13563 | 0.08086 | 1973 | 1 | 0.39538 | 0.12925 | |
| 1973 | 13867 | 0.12925 | 1974 | 1 | 0.62833 | 0.26122 | |
| 1974 | 14696 | 0.26122 | 1975 | 2 | 0.36702 | 0.38283 | |
| 1975 | 15460 | 0.38283 | 1976 | 3 | 0.45238 | 0.35912 | |
| 1976 | 15311 | 0.35912 | 1977 | 3 | 0.05778 | 0.40560 | |
| 1977 | 15603 | 0.40560 | 1978 | 3 | 0.83110 | 0.44667 | |
| 1978 | 15861 | 0.44667 | 1979 | 4 | 0.28212 | 0.59726 | Training data |
| 1979 | 16807 | 0.59726 | 1980 | 5 | 0.39200 | 0.61509 | |
| 1980 | 16919 | 0.61509 | 1981 | 5 | 0.48918 | 0.530560 | |
| 1981 | 16388 | 0.530560 | 1982 | 5 | 0.02845 | 0.37854 | |
| 1982 | 15433 | 0.37854 | 1983 | 3 | 0.38088 | 0.38872 | |
| 1983 | 15497 | 0.38872 | 1984 | 3 | 0.55037 | 0.33269 | |
| 1984 | 15145 | 0.33269 | 1985 | 2 | 0.84614 | 0.33556 | |
| 1985 | 15163 | 0.33556 | 1986 | 2 | 0.86535 | 0.46625 | |
| 1986 | 15984 | 0.46625 | 1987 | 4 | 0.46079 | 0.60553 | |
| 1987 | 16859 | 0.60553 | 1988 | 5 | 0.43712 | 0.81104 | |
| 1988 | 18150 | 0.81104 | 1989 | 6 | 0.55846 | 0.94157 | |
| 1989 | 18970 | 0.94157 | 1990 | 7 | 0.45431 | 0.99856 | Testing data |
| 1990 | 19328 | 0.99856 | 1991 | 7 | 0.97526 | 1 | |
| 1991 | 19337 | 1 | 1992 | 7 | 0.98835 | 0.92661 | |
| 1992 | 18876 | 0.92661 | - | - | - | - | |
| 1981 | | 16388 | 1992 | | | 18876 | |

## 4. Simulation study

To analyze the performance of the proposed method five real-time series are taken for the simulation study. First is the enrollment data of the University of Alabama from year 1971 to 1992 and the rest four are the daily closing prices Taiwan stock exchange from the year 2001 to 2004. In this study, we compared our proposed method with the methods presented by Aladag(2013), Bas et al.(2018), Panigrahi and Behera(2020) and Pattanayak et al.(2020). We also evaluate our proposed method by using multilayer perceptron (MLP) neural network to establish fuzzy logic relation and compare the results.

Aladag(2013) and Bas et al. (2018) proposed FTSF models for high order fuzzy time series. Aladag(2013) used a multiplicative neuron model to formulate the FLRs. Modified particle swarm optimization technique used to train the multiplicative neuron model for better forecasting. Bas et al.(2018) applied Pi-Sigma neural network for establishing FLRs and also used particle swarm optimization to optimize the weights of the Pi-Sigma neural network. Panigrahi and Behera(2020) implement three machine learning techniques namely deep belief

network(DBN), long short-term memory (LSTM), and support vector machine (SVM) for findings FLRs for the first time. We evaluate the FTSF-SVM model of Panigrahi and Behera(2020) for this study. Pattanayak et al. (2020) proposed a novel approach to divide UOD into the unequal length of intervals by FCM and used membership values with data to establish high order FLRs. We used python-based jupyter notebooks to implement all four comparison models and TensorFlow-Keras API for designing neural networks.

In order to implement our proposed method in jupyter notebook, first universe of discourse is defined then finds optimal number of clusters and get cluster centers by FCM algorithm using pre-defined scikit-fuzzy package in jupyter. By taking midpoints of cluster centers unequal lengths of intervals are calculated. Partition of UOD of all five time series into unequal intervals is given in Table 4. Then, the interval index number in which each datum of the time series falls is identified and membership value of the datum in that interval is calculated by the expression given by Eq. (8). The crisp time series is normalized by min-max scalar of scikit-learn package. After defining input-target pattern as defined in Table 2, the whole dataset is partitioned into training and testing data in the ratio of 80% and 20% respectively. The in-sample data (training data) is processed by 'SVM' module of scikit-learn and the input part of the out-sample data (testing data) is passed to trained SVM and predicted values are obtained. The obtained prediction is de-normalized by inverse min-max scalar of scikit-learn package.

For Multilayer perceptron (MLP) model TensorFlow-keras API is used to construct Multi-layer perceptron neural network. One hidden layer having the equal no. of neurons as in input layer is considered in every time series for fair comparison. All the graphs show in the study are plotted by matplotlib library of python programming package.

Table 3. Comparison of forecasted enrollment values of University of Alabama by different methods

| Year | Actual enrollment | Aladag et al. (2013) | Bas et al. (2018) | Panigrahi and Behera (2020) | Pattanayak et al. (2020) | Proposed method with MLP | Proposed method with SVM | |
|---|---|---|---|---|---|---|---|---|
| 1971 | 13055 | 13049 | 13049 | 13049 | 13055 | 13055 | 13055 | |
| 1972 | 13563 | 15049 | 15049 | 14049 | 13637 | 13826.48 | 14060.40 | |
| 1973 | 13867 | 15149 | 15149 | 14349 | 14120 | 14323.02 | 14497.26 | |
| 1974 | 14696 | 15149 | 15149 | 14549 | 14408 | 14620.31 | 14758.83 | |
| 1975 | 15460 | 15349 | 15349 | 15049 | 15195 | 14883.22 | 15017.41 | |
| 1976 | 15311 | 15549 | 15549 | 15549 | 15712 | 15588.55 | 15665.24 | |
| 1977 | 15603 | 15549 | 15549 | 15449 | 15635 | 15084.95 | 15222.17 | |
| 1978 | 15861 | 15649 | 15649 | 15649 | 15786 | 16071.89 | 16090.48 | Training data |
| 1979 | 16807 | 15649 | 15649 | 15749 | 15918 | 15967.65 | 16026.05 | |
| 1980 | 16919 | 15849 | 15849 | 16349 | 16406 | 16704.28 | 16701.41 | |
| 1981 | 16388 | 15949 | 15949 | 16449 | 16466 | 16828.30 | 16810.53 | |
| 1982 | 15433 | 15749 | 15749 | 16049 | 16190 | 16240.31 | 16293.21 | |
| 1983 | 15497 | 15549 | 15549 | 15549 | 15698 | 15497.30 | 15584.96 | |
| 1984 | 15145 | 15549 | 15549 | 15549 | 15731 | 15713.61 | 15775.27 | |
| 1985 | 15163 | 15549 | 15549 | 15349 | 15550 | 15494.69 | 15555.38 | |
| 1986 | 15984 | 15549 | 15549 | 15349 | 15559 | 15519.21 | 15576.95 | |
| 1987 | 16859 | 15649 | 15649 | 15849 | 15982 | 16195.28 | 16226.67 | |
| 1988 | 18150 | 15849 | 15849 | 16349 | 16433 | 16761.86 | 16752.08 | |
| 1989 | 18970 | 16149 | 16149 | 17149 | 17366 | 17513.11 | 17440.31 | Testing data |
| 1990 | 19328 | 16249 | 16249 | 17649 | 17967 | 17976.58 | 17875.35 | |
| 1991 | 19337 | 16349 | 16349 | 17849 | 18230 | 18641.56 | 18460.41 | |
| 1992 | 18876 | 16349 | 16349 | 17849 | 18236 | 18658.14 | 18474.99 | |
| RSME | | 2758.9 | 2758.9 | 1590.8 | 1342.3 | **1131.98** | **1211.08** | |
| SMAPE | | 15.603 | 15.603 | 8.6426 | 7.08 | **5.60** | **6.20** | |

## 5. Performance discussion

The performance measures RMSE and SMAPE of all the time series over test datasets is given in Table 5 and Table 6 respectively. Last two rows of both tables are performance measures of forecasted test data by proposed method using Multi layer Perceptron (MLP) and Support vector machine (SVM) for establishing relationship between input and target values. The best performance among all the models is highlighted by bold character. For the test data of student enrollment of University of Alabama RMSE and SMAPE by both proposed method is lesser than all four models i.e. Aladag et al. (2013), Bas et al.(2018), Panigrahi and Behara (2020) and Pattanayak et al. (2020). MLP gave better result than SVM on this data. In TAIEX 2001, last 49 closed prices are forecasted and observe that the proposed SVM approach outperforms overall methods. But in the TAIEX 2002 performance by the proposed methods perform worst than Panigrahi and Behera (2020) and Pattanayak et al. (2020). MLP approach of proposed method performs best among all methods in last two time series TAIEX 2003 and TAIEX 2004.

By the Table 5 and Table 6 we observe that out of five time series that we have used for this study four time series provide better result by proposed methods. Three time series namely Alabama Enrollment, TAIEX 2003 and TAIEX 2004 show less RMSE and SMAPE by computing MLP and one time series namely TAIEX 2002 best forecasted by SVM method. Fig. (1) to fig. (5) depicts the graph of all five actual time series, forecasted time series by proposed MLP and SVM approach. Actual time series is shown by blue color, proposed method by MLP and SVM are shown by green and orange color respectively. There is a dotted line present in the graphs which separate training and testing forecasted data. Training data is plotted before dotted line and testing data is plotted after dotted line. We can observe by the plotted graph in fig. (1) to fig. (5) that the proposed method fitted the time series very efficiently and gave a good forecast.

Table 4. Partitioning of UOD in unequal intervals of all five real time series

| Time Series | No. of clusters | Cluster centers | Partition UOD in unequal intervals |
|---|---|---|---|
| Albama Enrollment | 7 | 13573.95, 15130.14, 15448.22, 15885.33, 16825.08, 18190.36, 19125.23 | [13047, 14352.045]; [14352.045, 15289.18]; [15289.18, 15666.775]; [15666.775, 16355.205]; [16355.205, 17507.72]; [17507.72, 18657.795]; [18657.795, 19345] |
| TAIEX 2001 | 4 | 3946.44, 4499.63, 5194.15, 5637.06 | [3442, 4223.04]; [4223.04, 4846.89]; [4846.89, 5415.6]; [5415.6, 6108] |
| TAIEX 2002 | 4 | 4555.53, 4867.12, 5573.69, 6004.98 | [3846, 4711.33]; [4711.33, 5220.41]; [5220.41, 5789.34]; [5789.34, 6466] |
| TAIEX 2003 | 3 | 4,518.01, 5,144.84 5,796.52 | [4135, 4831.43]; [4831.43, 5,470.68]; [5470.68, 6146] |
| TAIEX 2004 | 2 | 5825.64, 6462.66 | [5312, 6144.15]; [6144.15, 7038] |

Table 5. RMSE on forecasted test data obtained by different methods

| Time Series | Aladag et al. (2013) | Bas et al. (2018) | Panigrahi and Behara (2020) | Pattanayak et al. (2020) | Proposed method with MLP | Proposed method with SVM |
|---|---|---|---|---|---|---|
| Alabama Enrollment | 2758.9 | 2758.9 | 1590.8 | 1342.3 | **1131.98** | 1211.08 |
| TAIEX 2001 | 536.76 | 475.02 | 121.07 | 119.38 | 120.98 | **117.93** |
| TAIEX 2002 | 530.59 | 533.90 | 71.22 | **67.96** | 89.94 | 82.52 |
| TAIEX 2003 | 511.03 | 148.91 | 56.88 | 63.37 | **53.56** | 66.73 |
| TAIEX 2004 | 336.74 | 364.47 | 54.41 | 54.39 | **53.92** | 55.25 |

Table 6. SMAPE on forecasted test data obtained by different methods

| Time Series | Aladag et al. | Bas et al. | Panigrahi and | Pattanayak et al. | Proposed | Proposed |

|  | (2013) | (2018) | Behera (2020) | (2020) | method with MLP | method with SVM |
|---|---|---|---|---|---|---|
| Albama Enrollment | 15.60 | 15.60 | 8.60 | 7.08 | **5.60** | 6.20 |
| TAIEX 2001 | 10.50 | 8.62 | 1.98 | 2.01 | 2.07 | **1.96** |
| TAIEX 2002 | 10.67 | 10.74 | 1.17 | **1.13** | 1.48 | 1.41 |
| TAIEX 2003 | 8.94 | 2.16 | 0.73 | 0.84 | **0.71** | 0.92 |
| TAIEX 2004 | 5.31 | 5.82 | 0.65 | 0.66 | **0.64** | 0.69 |

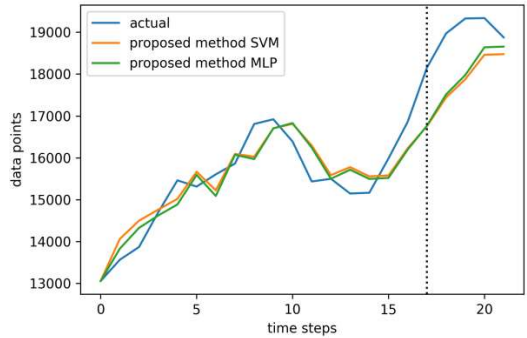

Fig 1. Actual and forecasted time series of enrollment data of University of Alabama

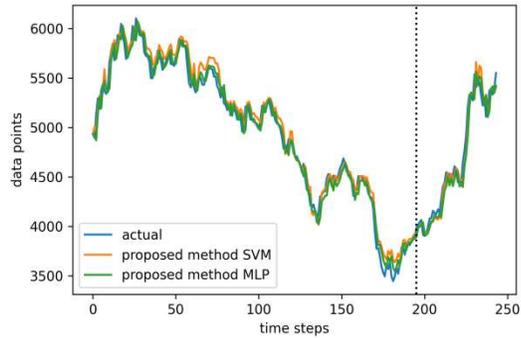

Fig 2. Actual and forcasted time series of TAIEX 2001

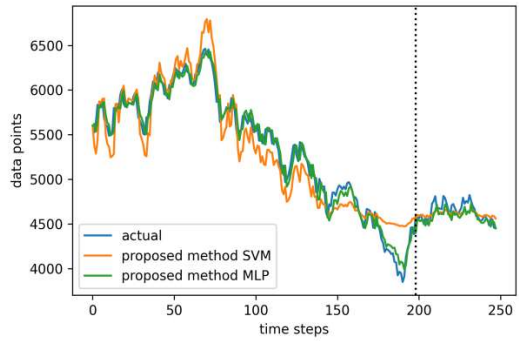

Fig 3. Actual and forcasted time series of TAIEX 2002

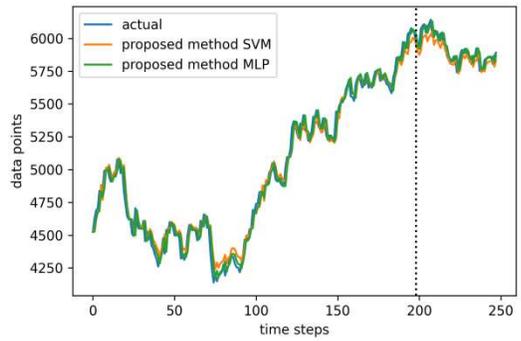

Fig 4. Actual and forcasted time series of TAIEX 2003

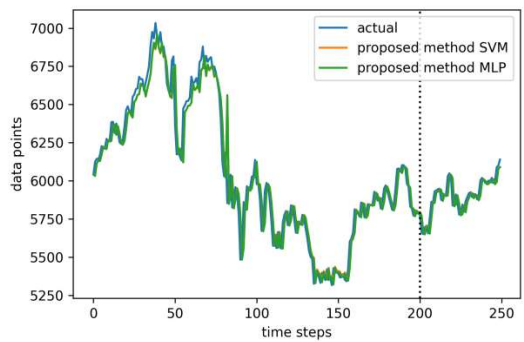

Fig 5. Actual and forcasted time series of TAIEX 2004

## (6) Conclusion

In this paper, a novel method is proposed for obtaining better fuzzy time series forecasting results. Rounding off range method with a suitable large step size is used to determine the optimal number of clusters for FCM. Cluster centers obtained by FCM are used to partition UOD into unequal length of intervals. This method of determining intervals help to find the appropriate number of intervals (NOIs) according to the range of time series and cluster centers ensure the density of the data into that particular interval. Time series is fuzzified into the interval obtained and interval index along with the membership or we can say position of the datapoint in the interval is used to forecast the future value. Proposed model is trained by support vector machine and multi layer perceptron neural network. The simulated study done on real time series verifies the success of the proposed model. The proposed forecasting method is first order time series forecasting because the features of the previous time step are taken to forecast the next value. So there is a future indication to examine this model on high order time series. Other methods of determining NOIs and partitioning of UODs can be explored to get more accurate prediction. Using optimization techniques to get most appropriate fuzzy logic rule can be beneficial to enhance the performance of the proposed model.